\documentclass[sn-basic,iicol]{sn-jnl}

\usepackage{graphicx}%
\usepackage{multirow}%
\usepackage{amsmath,amssymb,amsfonts}%
\usepackage{amsthm}%
\usepackage{mathrsfs}%
\usepackage[title]{appendix}%
\usepackage{xcolor}%
\usepackage{textcomp}%
\usepackage{manyfoot}%
\usepackage{booktabs}%
\usepackage{algorithm}%
\usepackage{algorithmicx}%
\usepackage{algpseudocode}%
\usepackage{listings}%

\usepackage[nolist,nohyperlinks]{acronym} 

\usepackage{xcolor}
\usepackage{pict2e}

\newsavebox{\ORCIDlogo}
\savebox{\ORCIDlogo}{%
\setlength{\unitlength}{\dimexpr 1em/256\relax}%
\begin{picture}(256,256)%
  \color[HTML]{A6CE39}\put(128,128){\circle*{256}}%
  \color{white}%
  \put(78.6,199.2){\circle*{20}}%
  \moveto(70.9,176,9)\lineto(86.3,176,9)\lineto(86.3,69.8)\lineto(70.9,69.8)%
  \closepath\fillpath%
  \moveto(108.9,176.9)\lineto(150.5,176.9)%
  \curveto(190.1,176.9)(207.5,148.6)(207.5 ,123.3)%
  \curveto(207.5,95,8)(186,69.7)(150.7,69.7)%
  \lineto(108.9,69.7)%
  \closepath\fillpath%
  \color[HTML]{A6CE39}%
  \moveto(124.3,83.6)\lineto(148.8,83.6)%
  \curveto(183.7,83.6)(191.7,110.1)(191.7,123.3)%
  \curveto(191.7,144.8)(178,163)(148,163)%
  \lineto(124.3,163)%
  \closepath\fillpath%
\end{picture}%
}
\newcommand\orcidicon[1]{\href{https://orcid.org/#1}{\usebox{\ORCIDlogo}}}

\begin{document}

\title[CausalOps - Towards an Industrial Lifecycle for Causal Probabilistic Graphical Models]{CausalOps - Towards an Industrial Lifecycle for Causal Probabilistic Graphical Models}


\author*[1]{\fnm{Robert} \sur{Maier}\orcidicon{0000-0002-4196-3263}}\email{robert.maier@oth-regensburg.de}

\author[2]{\fnm{Andreas} \sur{Schlattl}\orcidicon{0009-0008-1357-355X}}\email{Andreas.Schlattl@efs-techhub.com}
\equalcont{These authors contributed equally to this work.}

\author[2]{\fnm{Thomas} \sur{Guess}\orcidicon{0009-0004-3675-4952}}\email{Thomas.Guess@efs-techhub.com}
\equalcont{These authors contributed equally to this work.}

\author[1]{\fnm{Jürgen} \sur{Mottok}\orcidicon{0000-0002-7727-2448}}\email{juergen.mottok@othr.de}

\affil*[1]{\orgdiv{Software Engineering Laboratory for Safe and Secure Systems (LaS³)}, \orgname{OTH Regensburg University of Applied Sciences}, \orgaddress{\street{Seybothstraße 2}, \city{Regensburg}, \postcode{93053}, \state{BY}, \country{Germany}}}
\affil[2]{\orgdiv{Research and Development}, \orgname{e:fs TechHub GmbH}, \orgaddress{\street{Dr.-Ludwig-Kraus-Straße 6}, \city{Gaimersheim}, \postcode{85080}, \state{BY}, \country{Germany}}}



\abstract{Causal probabilistic graph-based models have gained widespread utility, enabling the modeling of cause-and-effect relationships across diverse domains. With their rising adoption in new areas, such as automotive system safety and machine learning, the need for an integrated lifecycle framework akin to DevOps and MLOps has emerged. Currently, a process reference for organizations interested in employing causal engineering is missing. This lack of guidance hinders the incorporation and maturation of causal methods in the context of real-life applications. To address this gap and foster widespread industrial adoption, we propose CausalOps, a novel lifecycle framework for causal model development and application. By defining key entities, dependencies, and intermediate artifacts generated during causal engineering, we establish a consistent vocabulary and workflow model. This work contextualizes causal model usage across different stages and stakeholders, outlining a holistic view of creating and maintaining them. CausalOps' aim is to drive the adoption of causal methods in practical applications within interested organizations and the causality community.}




\maketitle


\section{Introduction}\label{sec_intro}
In recent years, causal probabilistic graph-based models have emerged as powerful tools for illuminating cause-and-effect relationships. Their potential for bridging expert knowledge and data-driven analysis holds immense promise. However, integrating causal models into an organization's existing process landscape and their productive use in real-world applications remains a challenging endeavor. Currently, no best practices or guidelines exist that help organizations adopt causal engineering (i.e., using causal models to address business use cases) in a structured, manageable, or scalable way. The existing literature primarily focuses on isolated topics of causal models, overlooking the crucial need for a structured lifecycle framework that seamlessly aligns with lived practices. This gap in the research highlights the demand for a first draft of a comprehensive reference framework that presents a clear roadmap with distinct stages and entities, facilitating the smooth integration of causal models into established workflows. Such a framework fosters greater collaboration among stakeholders, empowers organizations to judge the required expertise for adoption, and harnesses the full potential of causal models in practical applications. 

Typically, reference frameworks emerge after a technology is already established. However, especially in the early stages of employing a methodology (similar to a low technology readiness level), guidance is essential to enable fast results and allow interested parties to adopt internal processes quickly. Therefore, the authors expect that providing such a framework early on will have the biggest impact on encouraging the widespread application of causal engineering.

Software development for instance has been on the rise for over seven decades. Along with a methodological maturation of \ac{sw} development, tools to support a product's lifecycle like integrated development environments, continuous integration pipelines, or code version management systems across a multitude of programming languages emerged. This led to the shortening of development time and helped to increase \ac{sw} quality and maturity~\citep{Survey_DevOpsQuality_2021}.
Today, \ac{sw} engineering is a multi-faceted, interdisciplinary branch of industry and one of the key technologies of the 21st century.

The perpetual lifecycle of a \ac{sw} product is often framed by the umbrella term DevOps~\citep{ISO_32675_DevOps,Survey_DevOpsQuality_2021,SW_DevOps_HighLevel_2016,SW_DevOps_Survey_2016}, combining the two main aspects: development and operations.

On the one hand, the \textit{Dev} aspect employs, among others, agile methods like Scrum~\citep{Book_Schwaber_Scrum_2001} to sufficiently address a highly dynamic environment (e.g., rapidly changing technology, requirements, or business models). On the other hand, the focus of the \textit{Ops} part is on customers and technical infrastructure. This includes providing and managing an already released product and a rapid, iterative, tightly coupled, interaction between development aspects and productive application. It covers publication, customer feedback, and error reporting and management.



DevOps allows structuring industrial processes on a large scale and therefore facilitates the rapid, high-quality development of \ac{sw} products. 
Similarly, related technologies like \ac{ml} are also passed a critical level of experience with regard to the development of algorithms and \ac{ml}-based products. Like in DevOps, MLOps~\citep{MicrosoftMLOps_Amershi_2019,MLOps_Taxonomy_2022,Survey_MLOps_2022} has been establishing a consistent vocabulary, taxonomy, and an overview of entities and distinct facets of product development.

\vspace{0.6em}
As a special kind of expert system~\citep{Book_Jackson_ExpertSystems_1999,Saibene_MedExpert_Systems_2021}, causal models are an emerging technology that allows building  an inferable model of a problem based on causal considerations.
Various methods to address cause and effect relations in a qualitative, quantitative, or probabilistic manner have been developed. 
The degree to which these approaches explicate modeled interactions as causal varies. A common denominator is the representation of causal influences via a graphical notation language as causal structures that constitute a simple way to represent a modeler's interpretation of causality among included variables. 

One of the most prominent implementation approaches for causal models, the probabilistic, graphical framework called \ac{bn}~\citep{Book_Koller_2009,Book_Jensen_BNs_2013}, has been developed for almost half a century and since then spread across a multitude of domains. 
Applications include manufacturing processes~\citep{ChemProduction_Zhu_2017,Lithium_Kirchhof_2020}, system reliability and dependability~\citep{CausalPerspecitve_Reliab_2020,Book_DBN_Reliability_2020}, aviation safety assessment~\citep{BN_AviationSafety_2019}, software engineering~\citep{CausalInf_SW_Siebert_2022}, econometrics~\citep{InferringCausalImpactTimeseries_2015,InfEng_HmundBareinb_2019}, epidemiology~\citep{Epidemiology_Robins_Hernan_2000,Epidem_Peterson_2014,EnvironmentalHealth_Bind_2019}, and \ac{ml}~\citep{DisentagleLearn_Bengio_2020,TowardsRepresLearn_Scholkopf_2021,CausalNeuralConn_2021,CausalML_Scholkopf_2022} among others.

Causal models can address these various business use cases due to their inherent ability to combine human expertise with data. 
Their use is often motivated by the necessity to predict the outcome of different (hypothetical) actions without the ability to examine them in a controlled environment. In many disciplines, it is not possible to measure the consequences of these potential actions as they might be either unfeasible (e.g., cost or resources) or ethically problematic (e.g, forcing one to take a drug). In these cases, causal models can be used to estimate potential effects, as they rely on a priori knowledge about a problem domain. 

Because of a diverse and clustered use of causal models, a taxonomy, a holistic view of lifecycle facets, and the identification of participating entities and their interdependencies is currently missing. Analogous to DevOps, a similar framework is needed to facilitate a unified view of causal engineering. 

To accelerate the maturation and encourage a widespread application of causal models, we propose a first draft of a lifecycle model called \textit{CausalOps}.

CausalOps can be coarsely split into two aspects, comprising the portmanteau\textemdash Causality guided development and operational use. Due to its intense use of software and \ac{ml} project aspects, it can be seen as a variant of DevOps and MLOps. 

Compared to MLOps, CausalOps focuses on causal models and an intense interdisciplinary communication of participating entities to maintain them. These models can also be seen as a special case of \ac{ml} models, to which concepts from MLOps may apply.

This article considers causal models as a product that is used to solve a problem at hand. This product is created based on a user's demands. The infrastructure, tooling, and associated processes remain constant for every newly created model instance. Therefore, instead of viewing the tailored creation of a causal model as the driving paradigm, we propose to view  a model's lifecycle as a recurring process that creates a customized product. This enables scaling processes, workforce, and infrastructure.

With this work, we provide:
\begin{itemize}
	\item a brief overview of causal engineering,
	\item a contextualization of our proposed lifecycle model within existing work,
	\item a definition of its facets,
	\item a definition of participating entities, 
	\item a definition of entity relations,
	\item a consistent taxonomy describing individual artifacts and lifecycle facets,
	\item and pointers to an integration in practice.
\end{itemize}

The focus will be exemplary on probabilistic, graphical models as defined by J. Pearl~\citep[Definition 7.1.6]{Book_Pearl_2009}. If not stated otherwise, the term \textit{causal model} used throughout this work refers to this framework. 

\vspace{0.6em}
The article is organized as follows: 

In Section~\ref{sec_context}, we provide additional background information on causality-guided models and related work that forms the foundation of this article. Section~\ref{sec_CausalOps} describes CausalOps in detail, outlining the overall concept, roles, and responsibilities of the workforce involved, and the intermediate artifacts that result from an application in practice. 

Next, two conceptual examples of the processual integration of the framework into existing workflows are given in Section~\ref{sec_application}. The approaches covered are expert-driven and integrative, as well as data-driven and complementary. Finally, we close the article with a summary of our key findings and pointers for future work.

An overview of relevant key vocabulary is provided in Appendix~\ref{subsec_vocabulary}. Readers are referred to Appendix~\ref{sec_Example} for a report of hands-on experience with a first application of the framework in the automotive domain, building on the expert-driven and integrative use case.

\section{Context}\label{sec_context}
This section differs from the traditional organization of an article in two ways. Firstly, we provide additional background information on causality-guided models. Secondly, as no directly related work besides general work on DevOps and MLOps exists, we present an overview of relevant publications that contextualize this paper. We expect that this approach may help readers unfamiliar with causal engineering better align our contribution with existing frameworks.

\subsection{Causality-guided Models}\label{sec_causalityGuidedModels}

\acp{dag} are a special and extensively used structure in graph-based approaches to causality. \acp{dag} consist of multiple nodes connected by directed edges and are required to be free of closed loops. They typically serve as a graphical notation language and facilitate causal inference~\citep{Book_Pearl_2009,Book_Scholkopf_2017}. 
Different extensions exist that link the informative and qualitative nature of \acp{dag} to data and probability theory. The most widely used probabilistic graphical framework is \acp{bn}~\citep{Book_Koller_2009,Book_Jensen_BNs_2013}. In \acp{bn} random variables and their conditional independence assumptions can be mapped  to a \ac{dag}. Variables become then its nodes, edges, and resulting substructures (i.e., 3-node clusters) correspond to independence assumptions (d-separation~\citep{DSeparation_Pearl_1990}).
As Dawid ~\citep{BewareOfDags_Dawid_2010} outlines, \acp{dag} and the problem space they model are not causal per default. Instead, this interpretation is given based on a model's context and constitutes a strong assumption about all parameters in a model.
Nonetheless, \acp{bn} allow an efficient combination of data, which provides probabilistic information, and domain knowledge which may justify the interpretation as causal.

A vast literature corpus exists that builds on the premise that conditional independencies between variables found in data can be algorithmically exploited and allow the construction of graphs based on observational data~\citep{Book_Neapolitan_algo,DyaLikeDags_2022}.
This field of research, called causal discovery, forms one of the two dominant approaches to model construction\textemdash data-driven and knowledge-driven.

Depending on the assumptions made, \acp{bn} can also be interpreted as a special instantiation of the more general framework of non-parametric \acp{sem}~\citep{Book_SEM_2015} called \acp{scm}~\citep{Book_Pearl_2009}.
Here, assignment-like expressions specify the interaction of causes to generate an effect.
These so-called structural equations imply a causal structure of the model in the form of a \ac{dag}~\citep{Book_Pearl_2009}. Moreover, \acp{scm} entail a set of models which can be implemented as \acp{bn}~\citep{OnPCH_Bareinboim}.

\acp{scm} are often considered as a causal interpretation of \acp{sem}. Both follow the premise that an interaction among variables can be attributed to a regime (i.e., some non-random mechanism) coupling them. This serves as the logical baseline to formulate causation on.
Causal structures can be used to represent different levels of causality, often termed as Pearl's \textit{ladder of causation}~\citep{Book_OfWhy_Pearl_2018} or \ac{pch}~\citep{OnPCH_Bareinboim}. \ac{pch} differentiates between the effects of purely correlation-based variable interactions (i.e., observed causality), active manipulation of a system (i.e., interventional causality), and reasoning about hypothetical, potential outcomes (i.e., counterfactual reasoning).
Each level can be framed in a graphical notation language and allows the qualitative investigation of an underlying problem domain by its representation as a causal structure.
\acp{bn} are typically used to model associational relationships. A corresponding graph therefore represents conditional independence assumptions that are expected to hold true in suitable (i.e., faithful~\citep{Book_Spirtes_Glymour_Scheines_1993,Book_Koller_2009,Unfaithfulness_Zhang_2008}) observational data.
Causal \acp{bn} (which are entailed by an \ac{scm}) allow representing active manipulations of variable interactions (in the form of edge manipulation). These graphical manipulations are formalized by a calculus of interventions (do-calculus~\citep{DoCalc_Pearl_1994}) that allows to transform a (hypothetical) action into an equivalent observational representation. \acp{swig}~\citep{SWIGS_Richardson_2013} extend the concept of a graphical representation of causality to allow the investigation of counterfactual questions for "what if" scenarios. 

Causal structures in the form of \acp{dag} are used as a framework to explicate and communicate assumptions and allow a qualitative evaluation of a modeled problem~\citep{Book_WhatIf_Hernan}. Depending on the specific framework (e.g., \ac{bn}, \ac{scm}, \ac{swig}), these assumptions can be supported by data and probability theory and allow an additional quantitative estimation (i.e., causal inference). 

Depending on their application, the above methodologies are subject to a context-specific definition of causality. Nonetheless, they are commonly used to model \textit{some} conditional relationship between variables in a defined problem space\textemdash either temporal precedence, logical association, or a regime-based generation of an outcome.

\begin{figure*}[t]
    \centering
    \includegraphics[width=\textwidth]{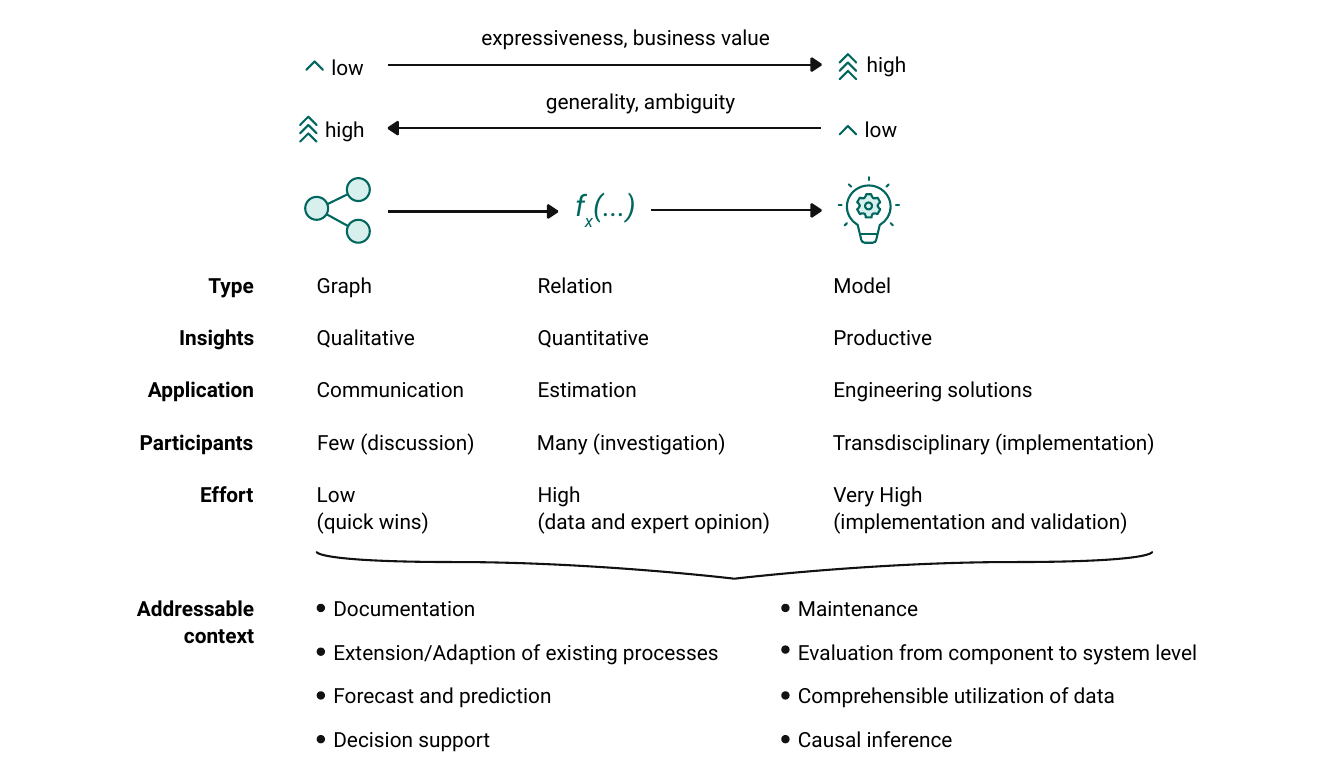}
    \caption{Overview of various application-relevant properties of causal graphs, relations, and models}
    \label{fig_app_dimensions}
\end{figure*}

Figure~\ref{fig_app_dimensions} provides an overview of various high-level aspects of qualitative models (i.e., causal graphs), causal relationships (mathematically expressed regimes), and causal probabilistic graphical models unifying both qualities. 

\subsection{Related Work}\label{sec_related_workd}
Causal models are an established methodology, yet little research is available that describes an end-to-end engineering workflow.

\textbf{Causal modeling:}
The development of causal models is typically decomposed into different stages. 
Boneh~\citep{KEBN_Boneh_2003} develops the concept of a knowledge engineering workflow that starts by specifying relevant variables (parameters) and consequentially adds structural compositions, and parameterization, and introduces testing and verification steps. Similar to Boehm's~\citep{SpiralModel_Boehm_1988} spiral model, these stages are recurring (if necessary) and allow an iterative approach to causal model creation. 
An associated engineering process detailing phase-specific considerations is provided by Korb and Nicholson~\citep[Chapter 3]{Book_BayArtIntelli_2010}. 

The general process of model construction can be separated into data-driven approaches and expert-driven modeling.

Literature on causal discovery (i.e., data-driven model creation) primarily focuses  on the mathematical preliminaries~\citep{DSeparation_Pearl_1990,Verma_Pearl_1990,LogicCausalModels_Pearl_1990,InterventionImplications_TianPearl_2002,Book_Spirtes_Glymour_Scheines_1993} and the development and evaluation of algorithms~\citep{Book_Neapolitan_algo,CausDiscovMethods_Glymour_2019,DyaLikeDags_2022,Survey_StrucutreLearn_2023}.

Nyberg et al. present a state-of-the-art expert-driven framework called Bayesian Argumentation via Delphi (BARD)~\citep{Bard_Tool_2021}. They combine the common knowledge engineering workflow of Korb and Nicholson with a Delphi~\citep{DelphiWithExperts_1963,Book_Delphi_1975} styled interviewing process, supported by an appropriate tooling landscape. Additionally, they define rudimentary roles (e.g., a moderator for group interviews) which take different responsibilities during model development. 

Maier et al.~\citep{Towards_CausalMBT_2022} link the iterative development and application of causal probabilistic models to the V-shaped process model of ISO~26262~\citep{ISO_26262}. They outline that a model may be used as part of an overarching engineering context, but do not provide details on the implications this has on model development or on participating entities (roles).

\vspace{0.6em}
\textbf{Workflow Concepts:} 
Published literature on causal modeling primarily covers the one-time, use-case-specific creation and subsequent application of causal models. These models are not considered as a product that may be continuously improved, like SW in \textit{*Ops} process models (where "*" acts as a placeholder). \textit{*Ops} as an umbrella term covers the domains of DevOps~\citep{SW_DevOps_HighLevel_2016,SW_DevOps_Survey_2016,SW_DevOps_Concepts_and_Challenges} (continuous development and integration of SW products) and MLOps~\citep{MicrosoftMLOps_Amershi_2019,MLOps_Taxonomy_2022,Survey_MLOps_2022,Survey_OperatMLModels_2022} (development and integration of data-intense \ac{ml}-based SW) among others (e.g., DevSecOps~\citep{DevSecOps_Akbar_2022}, ModelOps~\citep{ModelOps_Hummer_2019}, or DataOps~\citep{DataOpsApplication_Xu_2022}).

Alnafessah et al. discuss the implications of quality-aware tooling for each of the seven common stages of a DevOps cycle (plan, develop, verify, test, deploy, operate, and monitor)~\citep{Survey_DevOpsQuality_2021}. 
Kreuzberger et al.~\citep{MLOps_Kreuzberger_2022} extract seven common task-related roles (e.g., data engineer) that participate in an MLOps paradigm. Moreover, they show the interaction of these roles within an end-to-end MLOps workflow and identify rudimentary artifacts for different development stages.
Amaro et al.~\citep{Capabilities_Practices_DevOps_2022} investigate which capabilities (and in extension categories thereof like cultural or process aspects), practices (e.g., cross-team collaboration or artifact management) and their relationships shape current DevOps workflows.

Haakman et al.~\citep{AI_Models_Haakman_2021} conduct a case study that reviews current lifecycle models building on MLOps workflows. They identify missing aspects that challenge the application of ML-based SW in current practice. They highlight that typical stages like business understanding and problem modeling should be extended to include documentation, model evaluation, and monitoring activities.

Subramanya et al.~\citep{DevTOMLOps_Subramanya_2022} review connections between DevOps and MLOps, and discuss individual process steps (e.g., data preparation or performance monitoring) required to form a consistent MLOps pipeline.

\textit{*Ops} workflows are a multidisciplinary endeavor that relies on intense communication among different actors as well as an integration of multiple tooling paradigms.	
Honkanen et al.~\citep{MultiDiscTeamWorkMLOps_Honkanen_2022} provide an overview of current literature on different aspects of collaboration (e.g., team structures and roles, or effectiveness of communication) in MLOps workflows.
Hussmo et al.~\citep{AI_Models_Haakman_2021} discuss the role of engineering tools as boundary objects~\citep{BoundaryObjs_Star_1989,BoundaryObjsInstitutional_Star_1989} in transdisciplinary development. Heyn and Knauss~\citep{SCMasBoundObjs_Heyn_2022} propose to use \acp{scm} as suitable boundary objects in an MLOps workflow. They argue that \acp{scm} may be suitable to serve as a domain-independent framework to harmonize assumptions about an artificial intelligence-based system. Both publications build on the foundational works of Wohlrab et al.~\citep{BoundaryObjs_AgileEngineering_2019} and Carlile~\citep{BoundaryObjs_Carlile_2002,KnowlManage_Boundaries_Carlile_2004}. 
Wohlrab et al. discuss the role of artifacts in system engineering workflows, their management, and how boundary objects can be identified. 
Carlile discusses how knowledge can be efficiently managed across boundaries, and how boundary objects can serve as a flexible tool to facilitate  transdisciplinary communication and therefore improve product development.


\section{CausalOps}\label{sec_CausalOps}
The following section presents our proposed lifecycle model\textemdash \textbf{CausalOps}. It provides an overview of distinct stages, influencing aspects and required competencies, participating entities, their relations within individual lifecycle facets, and resulting artifacts that serve as intermediate products between them. 

\subsection{Proposed Framework}\label{sec_Framework}
Similar to SW in DevOps or \ac{ml}-products in MLOps, CausalOps can be split into different facets which cover phases like planning, model creation, publication, productive usage, and incremental improvement. This allows to define a continuos model lifecycle with distinct participating entities (roles), artifacts, and development and operation phases. 

\begin{figure}[t]
	\centering
	\includegraphics[width=\linewidth]{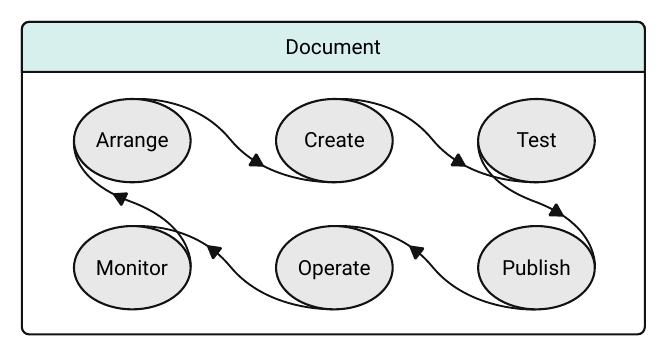} 
	\caption{High-level concept of the seven lifecycle facets of CausalOps}
	\label{fig_highlevel}
\end{figure}

The proposed elements are one way to specify a model's lifecycle. We are aware that each stage, role, or artifact can be defined at a different granularity. The level of detail may change based on an organization's resources (e.g., personnel, money), existing processes (e.g., production following a V-Model workflow), usage of causal models (e.g., for production, as documentation), or field of operation (e.g., model creator, model user). The presented distinctions made should therefore be seen as templates for a generic framework and a set of practices.
Figure~\ref{fig_highlevel} shows the proposed conceptual model of CausalOps.


As discussed above, a causal model's lifecycle is a multidisciplinary, multi-methodological, and multi-faceted endeavor. As a combination of different disciplines and aspects, a resulting lifecycle is characterized by an intensively coupled interaction of different facets: 

\begin{description}
	\item[\textbf{Arrange}] The initial phase of the causal model lifecycle. Based on the model's use case, context, and requirements a project plan for model development is created. Depending on the problem to be solved, experts and available knowledge is organized. The infrastructural resources and timelines are set up. 

	\item[\textbf{Create}] The model is created based on expert knowledge and data in accordance with the intended context of use. The actual development can be linear, iterative, or hybrid. Relevant tooling infrastructure  is created and maintained.
	
	\item[\textbf{Test}] The developed causal model is verified and validated against the existing requirements, data, and assumptions. The model is only tested with regard to technical considerations and not in the context of the intended productive model application.
	
	\item[\textbf{Publish}] The model and tooling infrastructure  is provided to the users. This includes the provision of a stable, executable configuration (parametrization, data, model version), customer training, and communication about model assumptions and limitations.

	\item[\textbf{Operate}] The model is in productive use. The model is verified and validated against the intended use case and employed in productive use on the user's side.
	
	\item[\textbf{Monitor}] Model performance, outputs, and insufficiencies and flaws are tracked. Data and knowledge bases are maintained and updated.
	
	\item[\textbf{Document}] The complete lifecycle of the model is documented. This includes facet-specific documentation in the form of artifacts, as well as generic documentation describing used methods, processes, and tools.
\end{description}

\subsection{Influencing Disciplines and Competency Clusters}
\begin{figure}[t]
	\centering
	\includegraphics[width=0.97\linewidth]{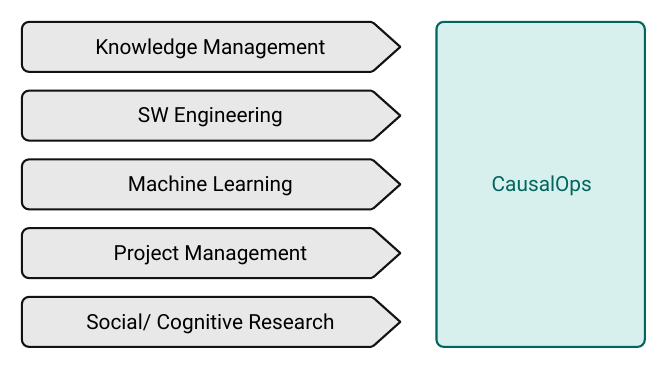}
    \caption{Disciplines influencing the theoretical foundation of CausalOps}
	\label{fig_influences}
\end{figure}

Causal models, their development, implementation, usage, and maintenance are an interplay of many domains of different competencies. Therefore, CausalOps is subject to the challenges of transdisciplinary engineering~\citep{TransdisEngi_2019}.
As outlined in section~\ref{sec_related_workd}, a successful lifecycle requires the management of resources, humans, and knowledge in the context of different boundaries~\citep{KnowlManage_Boundaries_Carlile_2004}. From a technical perspective, CausalOps relies on a combination of software engineering and \ac{ml}-elements. Figure~\ref{fig_influences} shows the main competency areas required: project management, knowledge management, social research, machine learning, and software engineering.

Project Management: CausalOps is designed to adopt an organization’s development strategy, such as linear (e.g., waterfall), incremental, agile (e.g., scrum), or hybrid approaches. Managing customer needs and allocating necessary resources are crucial for successful project completion, along with balancing the availability of domain experts with overarching development timelines.

Social Research: Intense communication is essential in CausalOps, and various techniques and methods need to be implemented throughout its lifecycle to achieve a high-quality iteration. Managing expectations, ensuring user/model interaction is intuitive; addressing biases during model construction (i.e., in expert-driven settings), and providing ongoing training for participating entities are key challenges.

Knowledge Management: Causal models represent a collection of knowledge, but original sources of this information can change. Tracking and documenting data and human expertise is essential, along with managing emerging new knowledge (e.g., tacit knowledge) during the model’s lifecycle. Preprocessing available sources of knowledge (e.g., technical documentation) and maintaining an up-to-date knowledge base are needed for the creation and correct application of the associated models. 

Machine Learning: Causal models are opaque in the sense that they can be interpreted and tested by verifiable assumptions~\citep{Grunbaum_QuantProbing_2022}. They still require verified, validated, versioned, and processable data. Known problems from \ac{ml}-intense applications, like distributional shifts and dependencies between training and productive use data, must be addressed.

Software Engineering: Causal models exist as mathematical specifications and executable software artifacts, both of which require dedicated efforts for building, implementing, and managing them. A software infrastructure for the continuous development and integration of causal models is not yet available. The application of CausalOps requires an intense effort from software engineers to transform published, theoretical groundwork into a user-friendly tooling infrastructure.

 \subsection{Participating Entities}
 The following listing defines the basic entities (roles) that participate in CausalOps. Alternative terms used in published literature (if available) are listed as an additional reference. In the context of existing processes and departments of an organization, these roles may be extended (e.g., quality management, requirements engineering, etc.) or combined (i.e., one individual may have multiple roles assigned).

 \begin{figure*}[t]
 	\centering
 	\includegraphics[width=\textwidth]{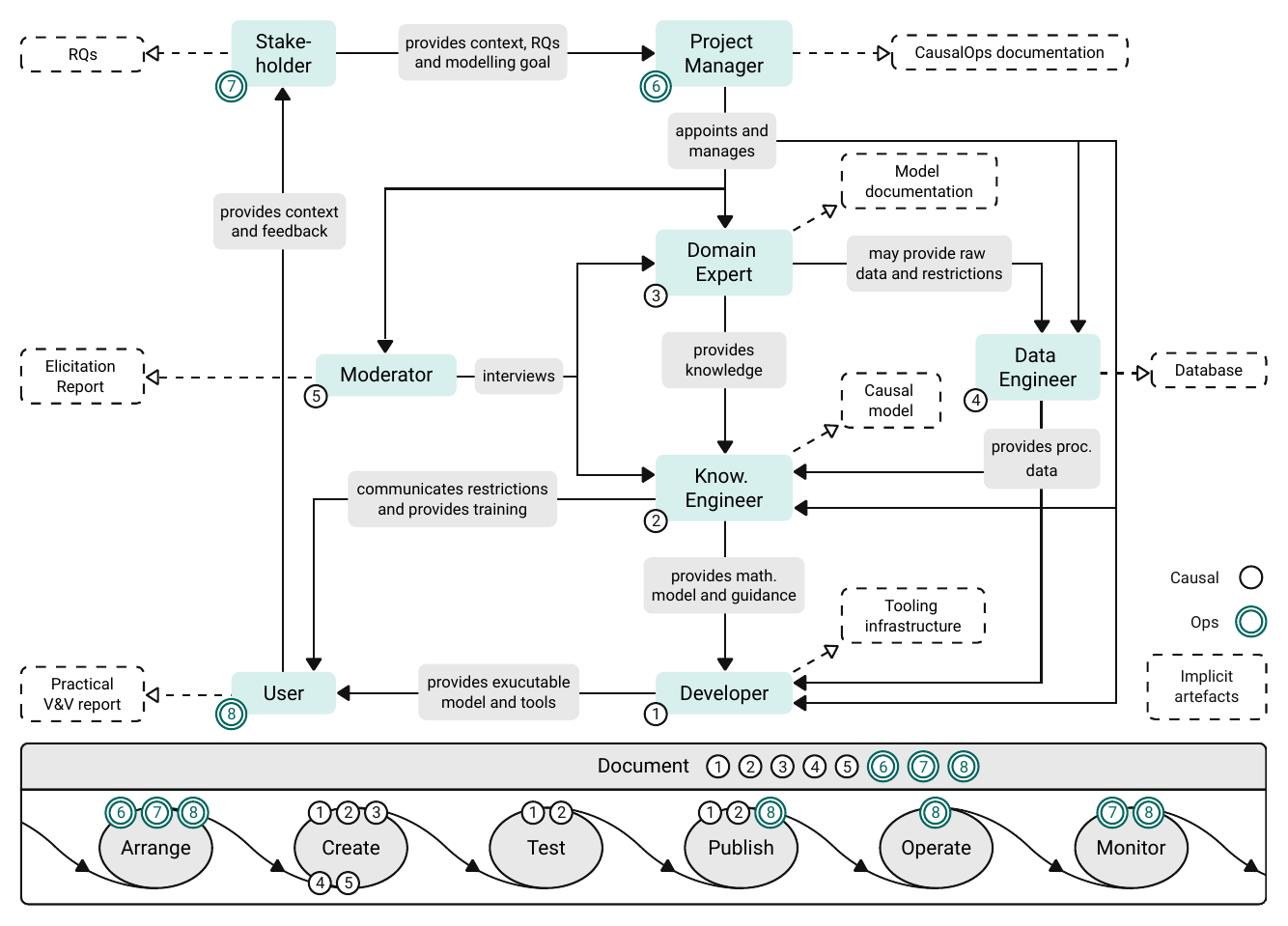}
 	\caption{Overview of relevant entities, their interdependencies, exemplary related artifacts, and their contribution in the individual lifecycle facets}
 	\label{fig_entities}
 \end{figure*}
 
 \begin{description}
 	\item[\textbf{Moderator}] A moderator (also called interviewer, facilitator, assessor, or receiver) is responsible for structuring and executing expert elicitation. They are responsible to manage discussions and communication among entities throughout the lifecycle with a focus on experts during elicitation rounds. 

 	\item[\textbf{Domain Expert}] A domain expert (also called expert, sharer, or analyst) serves as one source of information during causal model construction. They serve as a knowledge base due to their experience in a field that is deemed relevant for the current causal engineering effort. A domain expert shares knowledge during expert elicitation with regard to the requirements and the use case of model development.

 	\item[\textbf{Knowledge Engineer}] A knowledge engineer is responsible for the technical construction, population, verification, and validation of a causal model. They combine different, available sources of knowledge into a sound causal model.
 	
 	\item[\textbf{Data Engineer}] A data engineer is responsible to clean, structure, and preprocess data. They manage data that is required for model development and provide processed data to knowledge engineers and developers for incorporation into a causal model.
 	 	 	
 	\item[\textbf{Developer}] A developer is responsible for the implementation of a causal model as an executable software artifact. This includes providing and managing the software infrastructure to store, execute, test, and deliver causal models.
 	
 	\item[\textbf{Stakeholder}] A stakeholder serves as a triggering entity for model development. They provide context, requirements, and the use case for model development. Stakeholders act as a link between the operational and causal domains, by serving as initiators for model construction based on user demands.  
 	
 	\item[\textbf{Project Manager}] A project manager governs the complete model lifecycle and is responsible to manage model engineering. This includes time and resource management, team staffing, and customer management.

 	\item[\textbf{User}] A user is the primary operator of a causal model. They use causal inference to generate insights and apply them to solve the associated use case the model was developed for.

 \end{description}

Individual lifecycle members (i.e., project staff) may have multiple roles associated with them. Therefore, the outlined set of roles is to be seen independently of the actual number of project members.
Figure~\ref{fig_entities} provides an overview of entity interactions. Additionally, typical artifacts linked to each individual role are included, which are discussed in detail in Section~\ref{sec_BObjs_and_artifacts}.
The bottom part shows the mapping between the respective roles (entities) and the different facets of CausalOps in which they are primarily involved; this is further discussed below.


\textit{Stakeholders} are the initiators of causal model production and use. They provide a model's context, goals of the model application, and requirements for model content and quality expectations. Stakeholders can be directly involved  entities (e.g., customers, users) or a third party with aligning interests (e.g., authorities, government). 
\textit{Project managers} are the central communication and organizing authority. They are responsible to allocate resources (including experts with the required expertise) and organize the production process. Together with stakeholders, they serve as high-level participants who focus on generic project aspects.
Project managers are required to appoint and scout domain experts, moderators, knowledge engineers, data engineers, and developers. In a stable CausalOps environment, the biggest team fluctuations will be the \textit{domain experts}. They are selected based on a model's requirements for domain knowledge. Together with knowledge engineers, they participate in moderated elicitation processes. \textit{Knowledge Engineers} are experts in causal modeling. They convert cause and effect relations provided by domain experts and data into a suitable, mathematical framework (e.g., \acp{scm}), and are responsible for the technical specification of a  sound causal model. Throughout the lifecycle, and especially during expert elicitation rounds (i.e., in expert-driven model creation) \textit{moderators} structure the communication among experts. They are responsible to extract relevant knowledge from domain experts and knowledge engineers by interviewing them with regard to the modeling goals. As specialists in effective communication, they enable an efficient transfer of information  among all entities.
In data-driven model creation, moderators interview experts with regard to suitable data sources. The provided raw data is processed by data engineers into a utilizable representation for causal learning. \textit{Data engineers} manage available data and provide their characteristics and limitations to knowledge engineers.
After a stable causal model is specified in a mathematical framework, \textit{developers} transform it into an executable model. Developers are provided with relevant data and model descriptions by data engineers and knowledge engineers, respectively. Additionally, developers provide, manage and develop all relevant tools needed for the productive use of causal models. A subset of these tools, together with the executable model, is provided to the users. \textit{Users} are trained in the context of the model application (i.e., inference and result interpretation). They apply a causal model to solve the associated use case in practice. Based on model performance, initial requirements, and other restrictions, they collect and provide feedback for model improvement. Together with stakeholders and project managers, they represent the main actors of the "Ops" part of CausalOps. On the contrary, knowledge engineers, domain experts, data engineers, and moderators can be allocated primarily to the "causal development" part of CausalOps. Every entity is responsible to drive the documentation related to their respective role and the current lifecycle facet.

 With regard to the overall lifecycle model of CausalOps (see Figure~\ref{fig_highlevel}) each entity can be linked to at least one of its seven facets.
 During the arrangement stage, project managers, stakeholders, and users define the use case for which a causal model is required. For the actual development and implementation, close collaboration and intense communication among developers, knowledge engineers, domain experts, data engineers, and moderators is necessary. Once the causal model is well-defined and available as an executable, developers and knowledge engineers are able to test it against requirements and quality aspects (e.g., fit to data, plausibility). If the technical verification and validation activities are passed, the user is provided with the model, including specific training (if needed) and secondary information like restrictions.
 During the operating phase, the user actively uses model results to solve the use case at hand. Accompanying activities in the monitoring stage are managed primarily by the user and involved stakeholders. As an encompassing facet, documentation involves all parties at all times. 
 
\subsection{Boundary Objects and Artifacts}~\label{sec_BObjs_and_artifacts}
Boundary objects~\citep{BoundaryObjs_Star_1989,BoundaryObjs_Carlile_2002} are an essential tool to enhance the management of knowledge across different domains of expertise or disciplines. As \citep{BoundaryObjs_Features_2013,BoundaryObjs_AgileEngineering_2019} outline, boundary objects need to fulfill various criteria (e.g., modularity, concreteness). They serve as a key element to enable highly efficient, transdisciplinary communication by acting as a common format for knowledge exchange. With regard to CausalOps, causal models (depending on the actual mathematical framework used) qualify as boundary objects. As outlined above, CausalOps is comprised of seven different facets. Each stage is interdependent on existing information and simultaneously provides information for adjacent stages.  This information can be made available in the form of distinct artifacts. Some artifacts may also serve as boundary objects (e.g., the causal model itself or the context) to facilitate a participative and integrative engineering lifecycle~\citep{BoundaryObjs_AgileEngineering_2019}.

In the following, we provide a proposal of key elements for the outlined causal engineering lifecycle. It should be noted that this list is a result of reviewing existing literature (see section~\ref{sec_context}) and real-life experience. The intention is to highlight possible intermediate artifacts which are a by-product of causal model development and application. These artifacts act as living documentation and enable an efficient and transparent implementation of CausalOps.

\begin{figure*}[tbh]
	\centering
	\includegraphics[width=\textwidth]{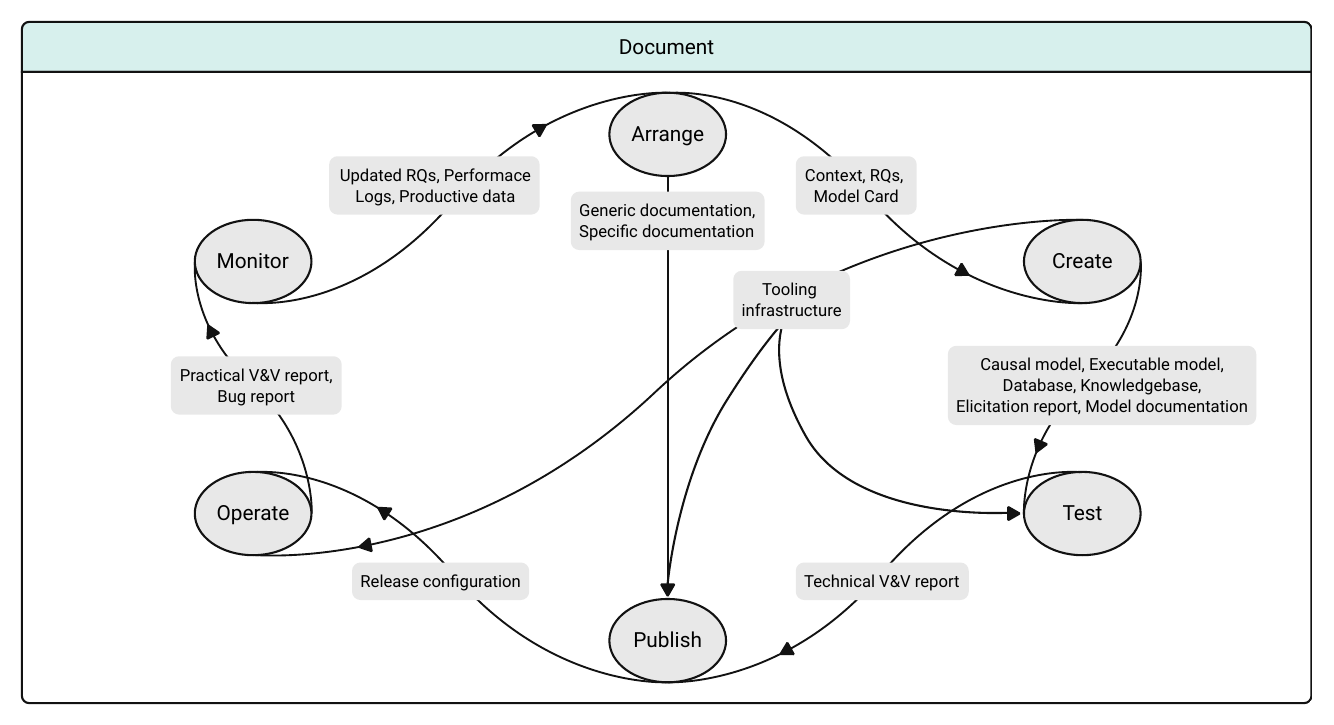}
	\caption{Overview of artifacts associated with individual lifecycle facets and their dissemination throughout CausalOps}
	\label{fig_artifacts}
\end{figure*}

Figure~\ref{fig_artifacts} shows where these artifacts are located in the CausalOps lifecycle. Note that each is available to all stages and therefore all entities of CausalOps once generated, and not only locally, as the depiction may imply.

The individual artifacts per lifecycle facet are provided below with a definition, followed by their intention (i.e., contribution to CausalOps), and associated entities which are primarily responsible for their maintenance.

\subsubsection{\textbf{Arrange}}
The arrangement stage is triggered by stakeholders. The main artifacts provided are a set of requirements and a context (e.g., as proposed by~\citep{CritAnalyAutomated_Neurohr_2021,GraspingCaus_DLR_2022}) of the model's intended application. The requirements define model-specific expectations, which can be technical (e.g., quality metrics) or abstract (e.g., able to address a specific subdomain of the use case). The context serves as an abstract definition of the use case and provides a restriction of the influences to be considered in a model (e.g., German highways as a restriction of the operational design domain in scenario-based vehicle testing). A model card~\citep{ModelCards_2019} is set up for the to-be-developed model at the project start. It serves as a high-level summary of all CausalOps stages, and their respective artifacts, and as a boundary object. During the arrangement stage, it is instantiated with available information (i.e., participants, context, and requirements (RQs)).
Model-independent, generic documentation containing causal engineering-specific descriptions serve as an overarching artifact. In practice, it may be a set of documents treated as a compiled knowledge database for the methods (e.g., Delphi method~\citep{Book_Delphi_1975}) and technologies (e.g., \ac{scm}) employed in CausalOps.

\begin{description}
	\item[\textbf{Context}] Set of high-level information, which provides the application domain and the rationals for the interpretation of causal mechanisms of a casual model. \leavevmode 
		\begin{description}
			\item[Intention:]  Defines the general conditions of model applicability, and guides the elicitation and knowledge extraction activities.
			\item[Entity:] 	   Stakeholder, User
		\end{description}
	
	\item[\textbf{Requirements}] Set of distinct, verifiable assumptions which outline the essence that should be expressed by a causal model. \leavevmode
		\begin{description}
			\item[Intention:]   Set the baseline to enable formal verification and validation, team staffing, and model development.
			\item[Entity:] 	   Stakeholder, User
		\end{description}
	
	\item[\textbf{Generic Documentation}] Compilation of knowledge about the employed tools and the methodology of causal modeling, causal models, and their inference. \leavevmode
		\begin{description}
			\item[Intention:] 	Summarizes current technical and scientific findings and documents the tooling landscape to enable the training of CausalOps entities with a focus on users.
			\item[Entity:] 	 	Project Manager, Knowledge Engineer, Developer
		\end{description}	

	\item[\textbf{Model Card}] Compact summary of all relevant meta-information with regard to a model. \leavevmode
		\begin{description}
			\item[Intention:]  Acts as a compact summary of all documentation efforts.
			\item[Entity:] 	   Project Manager
		\end{description}
\end{description}

\subsubsection{\textbf{Create}}
Based on the CausalOps strategy (expert-driven, data-driven, or hybrid), two bodies of information can be distinguished\textemdash data and knowledge.
All data that is directly used to learn or parametrize the causal model is part of a  database. All documents and other artifacts containing knowledge (e.g., ontologies~\citep{Survey_Ontologies_2021}) that contribute to the creation of a causal model are summarized by the umbrella term knowledge base. Together (database and knowledge base) they form the body of explicit information accumulated during model development and use. Parts of it may be provided by stakeholders and users, or during expert elicitation rounds. During these moderated sessions, tacit knowledge, trade-offs, and decisions are documented.
The development history of a model's construction is tracked in the form of an elicitation report. It enables the backtracking of model insights and possible insufficiencies down to alternative modeling decisions. Therefore, it serves as a central artifact to enable traceability. On a technical and model-specific level, model documentation detailing all parameters and elements including their meta-information (i.e., name, abbreviation, range of values, etc.) is used. It serves as a summary of the incorporated knowledge which is represented and expressed by a causal model.

The main artifacts of this stage are the mathematically specified causal model and its inferable representation as an executable model. The causal model itself is the central boundary object of CausalOps. In addition to the more technical model documentation, specific documentation outlining and summarizing model limitations (i.e., once the model is stable) with regard to the given context should be provided.

Depending on the degree of automation and extent of the software products used, the tooling infrastructure itself constitutes an artifact. Software tools like an inference engine~\citep{InfEngine_Pearl_2019,InfEng_HmundBareinb_2019} are essential, enabling components of CausalOps. This central module enables the transformation of a query (provided in a suitable format) into an estimate. Depending on their capabilities, the maturity and quality of a causal model may be subject to changes. Therefore, each version of a model needs to be associated with the individual tools (and their respective versions) used to allow a sound argumentation of insufficiencies or success in the model application.

\begin{description}
	\item[\textbf{Elicitation Report}]  Documentation of individual and group decisions documenting the creation history of a causal model.  \leavevmode
	\begin{description}
		\item[Intention:] Basis for preparing and conducting  elicitations, for collecting data, and for a seamless CausalOps documentation.
		\item[Entity:] Moderator
	\end{description}

	\item[\textbf{Database}] Collection of all relevant data which is associated with the causal model. \leavevmode
	\begin{description}
		\item[Intention:] 	Foundation for model parametrization in expert-driven workflows and model learning, and parametrization in data-driven and hybrid workflows.
		\item[Entity:] 	 	Domain Expert, Data Engineer
	\end{description}	
	
	\item[\textbf{Knowledgebase}] Collection of all relevant knowledge (e.g., documents, third-party artifacts) that is associated with the causal model. \leavevmode
	\begin{description}
		\item[Intention:] 	Serves as the secondary output of elicitation meetings and enables model construction and documentation.
		\item[Entity:] 	 	Domain Expert, Knowledge Engineer
	\end{description}	
		
	\item[\textbf{Model Documentation}]  Documentation of all model parameters and associated causal mechanisms, including their metainformation.  \leavevmode
		\begin{description}
			\item[Intention:]  Fundamental information about model elements to enable the implementation and application of a causal model.
			\item[Entity:] 	   Knowledge Engineer, Domain Expert
		\end{description}
	
	\item[\textbf{Causal Model}] The central artifact of CausalOps represented as a mathematical model. \leavevmode
		\begin{description}
			\item[Intention:] 	Main product.
			\item[Entity:] 	 	Knowledge Engineer
		\end{description}
	
	\item[\textbf{Executable Model}] Accompanying artifact that represents an inferable instantiation (e.g., software representation) of the underlying causal model. \leavevmode
	\begin{description}
		\item[Intention:] 	Basic artifact to operationalize (i.e., infer) and use a causal model.
		\item[Entity:] 	 	Knowledge Engineer, Developer
	\end{description}

	\item[\textbf{Specific Documentation}] Compilation of restrictions and characteristics of the causal model. \leavevmode
		\begin{description}
			\item[Intention:] 	Summarizes the current limits of the causal model and complements the given context.
			\item[Entity:] 	 	Knowledge Engineer
		\end{description}	
	
	\item[\textbf{Tooling Infrastructure}] Set of relevant tools to enable the creation, inference, testing, monitoring, and management of causal models. \leavevmode
	\begin{description}
		\item[Intention:] 	Enables the production and application of causal models.
		\item[Entity:] 	 	Developer
	\end{description}

\end{description}

\subsubsection{\textbf{Test}}
A technical verification and validation (V\&V) report is the central outcome of the test stage. It documents the technical insufficiencies of the developed model with regard to the context, requirements, and generic quality aspects. Therefore, it acts as the foundation to plan a release and publish a technical sound product.

\begin{description}
	\item[\textbf{Tech. V\&V report}]  Summary of all technical model evaluation activities.  \leavevmode
	\begin{description}
		\item[Intention:]  Serves as a certificate that the causal model is developed with regard to generic quality aspects and complies with the model-specific requirements.
		\item[Entity:] 	   Knowledge Engineer, Developer
	\end{description}
\end{description}

\subsubsection{\textbf{Publish}}
Depending on the overall iteration of the CausalOps lifecycle or inner iterations (e.g., model creation or software infrastructural changes) a release configuration needs to be defined. It allows consistent operational use of a causal model. This artifact enables the management of different models with the same context or a composition of a model based on different but semantically related sub-models.

\begin{description}
	\item[\textbf{Release Configuration}] Consistent description of compatible versions of the database, knowledge base, and other versioned documentation with regard to the current version of the causal model.  \leavevmode
	\begin{description}
		\item[Intention:]  Enables the traceability and versioning of a causal model.
		\item[Entity:] 	   Developer
	\end{description}

\end{description}

\subsubsection{\textbf{Operate}}
The primary artifact of this stage is a practical verification and validation report. While its technical counterpart only considers generic, use case-independent aspects, this report summarizes a model's capability to address the given context and use case. It is the outcome of the initial productive usage of a causal model after its publication. In addition, a summary of current errors encountered with the tooling infrastructure is given by a bug report. Both artifacts are unique and associated with a causal model release configuration.

\begin{description}
	\item[\textbf{Practical V\&V report}] Summary of all practical model evaluation activities with regard to the use case.  \leavevmode
	\begin{description}
		\item[Intention:]  Serves as an evaluation of a model's suitability to address all needs for productive use.
		\item[Entity:] 	   User
	\end{description}

	\item[\textbf{Bug report}] Summary of all known technical bugs and tool inconsistencies. \leavevmode
		\begin{description}
			\item[Intention:]  Serves as documentation to improve the associated tooling infrastructure of a causal model.
			\item[Entity:] 	   User, Developer
		\end{description}
	
\end{description}

\subsubsection{\textbf{Monitor}}
This stage generates artifacts that support an increment of the CausalOp lifecycle. Performance logs are used to track the operational efficiency of tools (e.g., runtime, memory) and model inference results (i.e., its outputs). Together with productive data (i.e., use case-specific data generated or collected in association with the productive application of a causal model), performance logs enable a continuous improvement of the tooling infrastructure (mainly the inference engine) and the model itself.
As a consequence of the operational performance of a causal model, a set of updated requirements can be defined. They are the baseline for a re-iteration of the CausalOps cycle and enable structured change management.

\begin{description}
	\item[\textbf{Performance Logs}] Collection of performance data including prediction accuracy and runtime measurements for employed tools and a causal model. \leavevmode
	\begin{description}
		\item[Intention:]  Summary of the technical monitoring activities which enable the continuous improvement of tools and of the causal model.
		\item[Entity:] 	   User, Developer
	\end{description}

	\item[\textbf{Productive Data}] Use case-related data that is generated or collected during the productive application of the causal model.  \leavevmode
		\begin{description}
			\item[Intention:]  May serve as an additional source of knowledge for model development, or as control data for model evaluation.
			\item[Entity:] 	   User
		\end{description}

	\item[\textbf{Updated Requirements}] Set of adjusted and new requirements derived from current model insufficiencies.  \leavevmode
	\begin{description}
		\item[Intention:]  Serves as the baseline for model improvement.
		\item[Entity:] 	   Stakeholder, User
	\end{description}

\end{description}

\subsubsection{\textbf{Document}}
Due to the multidisciplinary nature of CausalOps, generic process-oriented documentation is required. The CausalOps documentation serves as a collection of all model and tool-independent artifacts. It defines high-level activities and processes, terms and quality requirements, notations, and relations to and processes inherited from associated organization-specific development practices. Therefore, it structures the implementation of CausalOps as a methodology in a business and process context. Moreover, it defines activities and processes which will be inherited by CausalOps and their consequences on the artifact landscape. 

\begin{description}
	\item[\textbf{CausalOps Documentation}] Set of accompanying, model, and tool-independent documentation. It provides definitions of terms, processes, specifications of documentation activities, and quality requirements. It includes generic plans for project management (e.g., hierarchy, time, and resources), quality assurance, requirement management, verification and validation strategies, and responsibilities. \leavevmode
	\begin{description}
		\item[Intention:]  Enables an encompassing view of CausalOps in an organization's hierarchical development context and facilitates a consistent argumentation of all CausalOps activities to third parties.
		\item[Entity:] 	   Stakeholder, Project Manager
	\end{description}
		
\end{description}

\section{Integration Into Existing Workflows}\label{sec_application}
This section provides two conceptual examples of the integration of CausalOps into existing processes and the resulting consequences for the lifecycle model. This section intends to show how causal engineering can extend established workflows in an organization and the implications that such an approach entails.

\begin{figure*}[tbh]
	\centering
	\includegraphics[width=\textwidth]{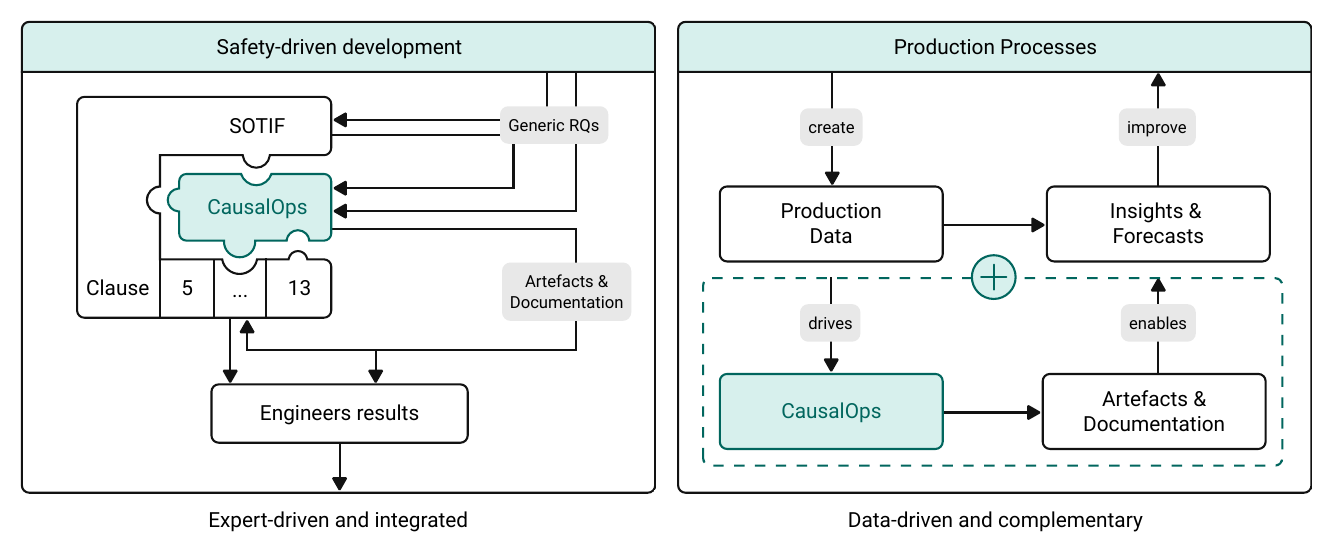}
	\caption{The conceptual model for the integration of CausalOps into existing workflows}
	\label{fig_application}
\end{figure*}

\subsection{Automotive System Safety}\label{sec_automotive_system_safety}
The development of modern vehicles is challenged by increasing system complexity. A recent approach besides technical safety (i.e., functional safety~\citep{ISO_26262}) is to ensure the \ac{sotif} of a vehicle. The normative regulation ISO~21448~\citep{ISO_21448_SOTIF} describes a set of practices and considerations which need to be fulfilled to label a system \textit{safe}. In the context of safety-driven system development, \ac{sotif} is only one aspect to evaluate a vehicle. Overarching process models like the V-Model structure the engineering landscape on a high level~\citep{ISO_26262}. They define generic requirements (e.g., documentation activities) which are subsequentially inherited and extended by integrated frameworks like \ac{sotif}.

In the case of \ac{sotif}, so-called \textit{clauses} define the stages of system evaluation, with the goal to provide a consistent and dependable safety argumentation. A cornerstone thereby is the identification and subsequent management of scenarios that may contribute to potentially hazardous events. As~\citep{Towards_CausalMBT_2022} outlines,  causal models can be used to model parts of a \ac{sut}. These models can also be used to identify potential triggering conditions that may be safety critical. Therefore, causal models and their development via CausalOps may be seen as an integrated methodology to support the \ac{sotif} evaluation of a \ac{sut}. 
As a consequence, global requirements of the enclosing engineering process and \ac{sotif} particularities are inherited by CausalOps.

These implications are collected and made available in the CausalOps documentation (see Section~\ref{sec_BObjs_and_artifacts}) and affect all documentation activities and artifacts.
Moreover, it may result in an expert-driven approach as the outer process landscape (i.e., V-Model) heavily relies on domain experts. Therefore, CausalOps-specific artifacts need to be linked to the individual engineering activities of \ac{sotif} (i.e., its clauses). This is needed to guarantee a consistent argumentation of causal model results in the context of a safety argumentation.

CausalOps may extend the set of methodologies used to evaluate an autonomous system. This has direct implications on the internals of CausalOps:

\textbf{Affected Artifacts:} \hspace{0.75em}
The \textit{context, use case, and requirements} are extended by the \ac{sotif}-specific goals outlined in the normative clauses of ISO~21448, and derived from the individual \ac{sut}. The \textit{generic documentation} is similarly affected by the quality and documentation guidelines and processes inherited from the V-model and ISO~26262.
Due to causal models being only one of the applied methods to ensure system safety, the results of already employed approaches (e.g., hazard analysis and risk assessments)  define a large part of the available \textit{knowledge base}. Moreover, the engineering and real-life data of a \ac{sut} constitute  the core of the \textit{database}.

\textbf{Affected Entities:} \hspace{0.75em}
Due to the broader context of guaranteeing system safety, \textit{stakeholders} may be the head of the safety department inside the practicing organization as well as government institutions.
Participating safety engineers become the \textit{users} who apply causal models to generate safety-relevant insights. During model construction, system and safety engineers act as \textit{domain experts} to build a suitable causal model.

The left part of Figure~\ref{fig_application} shows the abstract interaction between global safety processes, the \ac{sotif} framework, and CausalOps as an integrated module.
An additional example is given in Appendix~\ref{sec_Example}, providing hands-on experience of an application of CausalOps in the automotive domain.


\subsection{Manufacturing Industry}
In many automated manufacturing processes, failure detection and forecast are essential aspects~\citep{Lithium_Kirchhof_2020,ChemProduction_Zhu_2017}. Typical production lines are highly complex and are comprised of a vast variety of sensors and actuators. Both are densely monitored and resulting data is logged. These rich databases allow a data-driven use of causal models and therefore the application of CausalOps as a suitable methodology to improve production processes. In the simplest form, only the available data is used to drive model development. This is enabled by a variety of algorithms for causal discovery and subsequent parametrization. The focus of CausalOps internals shifts to data engineers who preprocess data and developers who establish and improve suitable end-to-end tools~\citep{Grunbaum_QuantProbing_2022}. 
The resulting artifacts (primarily the executable model)  can then be incorporated into the manufacturing process\textemdash either automatically to support production infrastructure and control processes or to support maintainers by providing estimates for faulty components. The resulting forecasts and insights act as drivers for a continuous improvement of all associated production processes. For the stand-alone application of CauaslOps this may have the following implications among others:

\textbf{Affected Artifacts:} \hspace{0.75em}
The \textit{executable model} may be fully automatically created by a suitable \textit{tooling infrastructure}, which takes processed production data as input. A causal model including an accompanying \textit{model documentation} may then be a result derived from an associated \textit{database}. Depending on the degree of automation, elicitation processes may only focus on the selection of data. A subsequent \textit{elicitation report} may therefore only describe the selection criteria for the algorithms used throughout the tool chain. The tooling infrastructure focuses on data pipelines, which provide learning algorithms with preprocessed input. 
The inference of an \textit{executable model} can be automated based on the specific integration into production. \textit{Performance logs} and productive data become a driving aspect for a continuous improvement of the causal model and extend the \textit{database} used for model development.

\textbf{Affected Entities:} \hspace{0.75em}
The driving entity for a stand-alone and data-driven application of CausalOps are \textit{developers} and \textit{data engineers}. They provide the technical infrastructure to automate causal model creation and use. 
\textit{Domain experts} in the form of manufacturing engineers and production workers together with \textit{knowledge engineers} only provide high-level input for a suitable selection of algorithms and data. 
Furthermore, they are the driving entities to provide feedback during the test and operation stages. 
\textit{Users} depend on the type of integration of causal models into the production pipeline. The primary users may be system engineers and maintainers in the context of fault diagnosis.
The right part of Figure~\ref{fig_application} shows the integration of CausalOps as a stand-alone methodology to continuously improve manufacturing processes.


\section{Conclusion}\label{sec_conclusion}
In this paper, we showed a first draft of a lifecycle model for the continuous development and maintenance of causal models, called CausalOps We presented influencing concepts of related domains, established frameworks like DevOps and MLOps, and transdisciplinary engineering, which form the theoretical foundation of this work.

Our main contribution is providing a basic vocabulary, a conceptual lifecycle model for causal engineering, an overview of involved entities, their roles within different facets of our framework, and resulting work products. Finally, we presented pointers to an application in practice.

For a widespread application and subsequent industrialization of causal engineering, the challenges outlined in this article remain. These include developing automated model verification and validation techniques, software tools to support expert-driven development, or a process framework to enable and guide organizations in implementing causal engineering at scale. 

Technical aspects of causal models, like the tractable calculation of counterfactual queries, and practical challenges like the efficient training and education of an organization’s diverse workforce, or the procedural integration of a new method in existing workflows, are areas of ongoing research.

Causal probabilistic graph-based models can be used as powerful tools to capture and understand complex relationships, but the challenge of developing best practices and reliable guidelines on how to employ them efficiently and effectively remains.

We hope CausalOps can serve as a conceptual starting point to harmonize the development and application of causal models across different domains and act as a catalyst for the rapid maturation of causal engineering.

\backmatter





\bmhead{Acknowledgments}
We thank Clara Cullmann-Petroll and Katharina Romanow from "e:fs TechHub GmbH", and Ascan-Olaf Lutteroth from "Volkswagen Commercial Vehicles" for their invaluable feedback. Moreover, we thank Daniel Ebenhöch from "e:fs TechHub GmbH", Tilo Linz from "imbus AG", and their respective team members for support, valuable insights, and discussions.

\section*{Declarations}

\bmhead{Contributions} All authors contributed to the study conception and design. Material preparation, data collection and analysis were performed by Robert Maier and Jürgen Mottok. The first draft of the manuscript was written by Robert Maier and all authors commented on previous versions of the manuscript. The Appendices were written by Robert Maier, Andreas Schlattl, and Thomas Guess. All authors read and approved the final manuscript.

\bmhead{Funding} This work is partially supported by the 'Bavarian State Ministry of Economic Affairs, Regional Development and Energy' (STMWI) through the granting of the funding project HolmeS³ (FKZ: DIK0173/03).

\bmhead{Competing interests} The authors declare that they have no competing interests.

\bmhead{Ethics approval} Not applicable.

\bmhead{Consent to participate} Not applicable.

\bmhead{Consent for publication}  Not applicable.

\bmhead{Availability of data and materials}  Not applicable.

\bmhead{Code availability} Not applicable.

\bmhead{Open Access} 
This article is licensed under a Creative Commons Attribution 4.0 International License, which permits use, sharing, adaptation, distribution and reproduction in any medium or format, as long as you give appropriate credit to the original author(s) and the source, provide a link to the Creative Commons licence, and indicate if changes were made. The images or other third party material in this article are included in the article’s Creative Commons licence, unless indicated otherwise in a credit line to the material. If material is not included in the article’s Creative Commons licence and your intended use is not permitted by statutory regulation or exceeds the permitted use, you will need to obtain permission directly from the copyright holder. To view a copy of this licence, visit \url{http://creativecommons.org/licenses/by/4.0/}.


\bigskip
\begin{flushleft}%

\bigskip\noindent

\end{flushleft}

\begin{appendices}

\section{Vocabulary}\label{subsec_vocabulary}

The following set of definitions forms the basic vocabulary relevant for this article and is based on existing literature (e.g,~\citep{Book_Pearl_2009,Book_Koller_2009,Book_Scholkopf_2017,Book_WhatIf_Hernan}):

\begin{description}
	\item[\textbf{Node}] A node represents a distinct, fundamental element of a graph. It is the representative of a cause or effect depending on its relative position in the graph.  
 
	\item[\textbf{Edge}] An edge is a fundamental element of a graph typically represented as a (directed) line that connects two nodes. It indicates that a distinct regime couples these nodes and symbolizes an underlying association (e.g., correlation or causation) between them. 
	
	\item[\textbf{Causal Mechanism}] A causal mechanism specifies the interaction between one or more causes and a resulting effect. Depending on the framework used it may be formalized as a regime or in the form of (structural) assignments (i.e., in \acp{scm}).
	
	\item[\textbf{Causal Structure}] A causal structure (also called a causal graph or diagram) is the graphically (implied) representation of a set of causal mechanisms. A causal structure is often visualized as a directed, acyclic graph where cause and effect entities form the set of nodes.
	
	\item[\textbf{Causal Model}] A causal model consists of a set of causal mechanisms and their implied causal structure, and defines the mathematically  formalized instantiation of a modeled problem. For the formal definition of a causal model as considered in this article see~\citep[Definition 7.1.6]{Book_Pearl_2009}.  
	
	\item[\textbf{Executable Model}] An executable model is the implementation of a causal model as a software product or in a suitable exchange format. It allows a tool-based utilization of a causal model.
	
	\item[\textbf{Query}] A query describes a question relevant to a modeled problem domain. It is formalized in a suitable query language and is valid for a specific causal model.
	
	\item[\textbf{Estimate}] An estimate is the (numeric) answer to a query derived from a causal model.
	
	\item[\textbf{Inference}] Inference describes the calculation of an estimate for a given query based on a causal model.
	
	\item[\textbf{Context}] A context defines a set of high-level information for a causal model's applicability and additional a priori known expectations about the problem domain. It defines the rationale and background of a model.
	
	\item[\textbf{Requirements}] The requirements (RQs) define a set of verifiable specifications that need to be expressed by a causal model as well as strategies to evaluate them (e.g., validation strategy, consistency checks). Requirements may include metrics and thresholds to assess a model's quality with regard to data.
	
	\item[\textbf{Use case}] A use case describes the intended use and goals of a model's application. It makes use of the context and defines desired insights from the model used to solve the problem at hand.
	
	\item[\textbf{Causal modeling}] Causal modeling describes the activity of causal model creation. It can be either data-driven (i.e., reliant on algorithms and  existing data of the problem domain), knowledge-driven (i.e., reliant on human expertise), or hybrid (i.e., a combination of algorithms and domain expert knowledge).

     \item[\textbf{Causal engineering}] Causal engineering describes the process of developing and utilizing causal probabilistic graphical models to address a business use case with different stakeholders involved. CausalOps is an associated reference lifecycle framework that formalizes entities and artefacts involved in causal engineering.
	
\end{description}

\section{Automotive Example}\label{sec_Example}
This section presents practical experiences within the research project \textit{HolmeS³} funded by the "Bavarian State Ministry of Economic Affairs, Regional Development and Energy" (STMWI). 

In order to illustrate an iteration of CausalOps we refer to the integration example given in Section~\ref{sec_automotive_system_safety} and substantiate it via an exemplary validation of a hypothetical \ac{adas} in the context of the \ac{sotif}. We consider the use case of ensuring the safety of an \ac{eba} as a simple, comprehensible, high-level example throughout this section. A comprehensive discussion of an \ac{eba} is beyond the scope of this work. Interested readers are referred to the technical standard ISO~22839~\citep{ISO_22839} for extensive information about requirements for a deployable \ac{eba}.

Using causal models as a virtual proving ground, an evaluation of an \ac{eba} might be addressed by CausalOps in the following way:

 \textbf{Arrange:} An automotive manufacturer requesting the validation of an \ac{eba} takes the role of a \textit{stakeholder}. During the \textit{arrange} phase, a \textit{project manager} is entrusted by the \textit{stakeholder} with the creation of a causal model and the derivation of safety-relevant insights (i.e., the \textit{use case}).
 
 Firstly, the \textit{project manager} collects the \textit{stakeholder's} \textit{requirements} and defines the automotive system's \textit{context} (e.g., German highways only). For an \ac{eba}, one \textit{requirement} may be that under certain circumstances (i.e., within its specifications) the system has to guarantee that crashes are sufficiently mitigated. Secondly, the \textit{project manager} organizes consultations between \textit{domain experts}, the designated \textit{moderator}, and the \textit{knowledge engineers}. In order to model the \ac{eba} behavior  correctly, \textit{domain experts} from the entire functional chain (i.e., spanning sense, plan, and act activities of the \ac{eba}) are required. Sensor experts may contribute by specifying how weather conditions influence the detection of relevant objects (e.g., pedestrians or cars) ahead of the equipped vehicle, while engineers focusing on motion planning enrich the model by providing causal relations between factors like the relative speed between the equipped vehicle and a detected object, or the activation thresholds of the system. Moreover, physicists may provide information about the influence of the road's friction on the breaking behavior of the \ac{eba}-equipped vehicle.
 A \textit{moderator} provides a collaborative working environment including tools like real or virtual whiteboards, or dedicated software required for a knowledge elicitation process (e.g., ~\citep{Bard_Tool_2021}). Agendas for elicitation meetings are defined in order to specify the relevant nodes, edges, mechanisms, and probability distributions for the causal model.

\textbf{Create:} The derivation, collection, and documentation of knowledge are part of the \textit{create} phase. 
In each of the following steps, the \textit{knowledge engineers} have to decide whether to use data or expert knowledge for the creation of the causal model. For the running example, an expert-driven approach to derive a rudimentary model is followed first.
Here, the initial step may focus on extracting the nodes, and thereby the causal variables of a model from domain knowledge. Possible questions for an initial expert interview round could be: 
\begin{itemize}
    \item What are possible hazardous behaviors  of an \ac{eba}?
    \item What are potential causes for known functional insufficiencies?
\end{itemize}
As a result, the \textit{domain experts} may suggest that rain and fog are essential environmental  parameters that affect the triggering conditions of an \ac{eba} and should be included in the set of relevant parameters.
Next, edges between available nodes can be defined either by causal discovery algorithms~\citep{DyaLikeDags_2022} or by \textit{domain experts}. 
Finally, probability distributions and causal mechanisms are specified and used to complete an initial version of the causal model.

We experienced that these interviews are highly fruitful, as group discussions (i.e., as part of a moderated elicitation process) enable one to address relevant parameter constellations that might otherwise have been missed early on (i.e., due to the split into different business units and their respective competency areas in practice). It should be noted that finding relevant nodes is a creative process. Therefore, a moderator needs to be familiar with group creativity techniques like brainstorming, and able to re-align discussions if required.
The participation of multiple experts from one domain, and also across domains, should be pursued in these interviews, as model creation might be significantly improved. In subsequent, iterative model-refinement sessions dealing with the specification of edges, causal mechanisms, and probability distributions, single-expert interviews proved valuable.

Furthermore, an early grouping of experts with similar expertise (e.g, covering various, relevant sensing technologies like RADAR or LiDAR) to specify parts of the functional chain (e.g., sensor data fusion, trajectory planning) helped to identify missing expertise for the \ac{eba} use case for subsequent modeling activities. 

The implicit history of the models' creation detailed above, including  participating \textit{domain experts}, questions asked, rationales and datasets given, and algorithms used, is part of the \textit{elicitation report}. Provided insights, discussion protocols, and pointers to relevant literature are collected as part of the \textit{knowledge base}. 
Storing the causal relations in a \ac{dag} exchange format facilitates the transformation from a causal model to an \textit{executable model}. Moreover, this allows for visualization of the current graphical structure of the causal model and allows a utilization by established graph processing software such as GraphViz\footnote{\url{https://graphviz.org}}, NetworkX\footnote{\url{https://networkx.org}}. 

\textbf{Test:} If proper tooling is available, the test stage can be conducted automatically, similar to continuous integration pipelines in the field of SW development. Tests for the \textit{executable model} can be configured in the manner of \textit{unit tests} by providing verifiable causal assumptions~\citep{Grunbaum_QuantProbing_2022}.  In this phase, \textit{developers} and \textit{knowledge engineers} typically collaborate to transform the causal expectations provided by \textit{domain experts} (e.g., intense rain increases crash likelihood)  into automatically executable tests. A model is considered stable if all requirements are met and all tests evaluating technical qualities are passed. 

\textbf{Publish:}
In the case of the running example, the outer \ac{sut} development milestones may influence  the iterations of the CausalOps lifecycle. The \textit{stakeholder} provides due dates and maturity requirements of intermediate deliverables. Publication in this context can be considered as an ongoing release. In the example above, a release covering available artifacts (e.g., \textit{elicitation protocols, knowledge base}, and the current version of the \textit{executable model}) may be requested after every expert elicitation meeting. For instance, following a first interview round, the \textit{stakeholder} may be provided with a list of environmental conditions such as rain and fog which, by expert definition, may contribute to a potentially hazardous behavior of the \ac{eba} system. Subsequently, after the second interview round, combinations of parameters representing causal sub-structures (and therefore autonomous mechanisms) in the causal graph can be provided to the \textit{stakeholder}. Moreover, in combination with the intermediate causal model, estimates based on simple  queries may provide first insights into safety-critical phenomena~\citep{GraspingCaus_DLR_2022}. For instance, one such result may show how rain and fog influence the distance at which the \ac{eba} is activated, which in turn influences the likelihood of a collision. 

\textbf{Operate:} 
The \textit{operate} stage is used to extract safety-relevant parameter constellations from the causal model (use case). This can be done by formulating queries that can be interpreted by specialized software (i.e., a causal inference engine).  For the running example, a potential query could be "What is the probability of a collision if the values of fog and rain are simultaneously changed to a certain level?". The result of multiple such queries may be aggregated and provides indicators for potential triggering conditions resulting in unsuccessful emergency braking.
Moreover, combinations of parameter values that constitute a potentially challenging test case (e.g., a corner case) for the \ac{eba} can be extracted from the model~\citep{Towards_CausalMBT_2022}.

\textbf{Monitor:} The monitor phase involves ongoing performance monitoring of the \ac{eba} causal model, using collected data and knowledge to identify areas for improvement. Inference results of the causal model may show that, depending on the daytime, model estimates diverge from the results of simulation tests. into distance at activation to the causal model. 
To address this in an iteration of CausalOps, these insights are included either in the \textit{practical V \& V report} or in the \textit{bug report} if results can be attributed to inference software insufficiencies.

\textbf{Document:} The \textit{generic documentation} serves as a reference for all development and application efforts. It enables involved organizations to implement a consistent and systematic approach to \ac{adas} system development based on causal models and allows an attributable integration into existing processes (e.g. \ac{sotif}, or ASPICE\footnote{\url{https://www.automotivespice.com/}}). 
For the running example, the \textit{generic documentation} needs to outline how these accompanying guidelines are addressed by causal engineering and by CausalOps in particular. All of the activities above are summarized and can be used to update an initial set-up \textit{model card}, which can be used as a starting point for a potential re-iteration of CausalOps.




\end{appendices}

\begin{acronym}[SEPSEPSEP]
 	\acro{ml}[ML]{Machine Learning}
 	\acro{sw}[SW]{Software}
 	\acro{swebok}[SWEBOK]{Software Engineering Body of Knowledge}
 	\acro{sotif}[SOTIF]{Safety of the Intended Functionality}
 	\acro{ft}[FT]{Fault Tree}
 	\acro{et}[ET]{Event Tree}
 	\acro{fcm}[FCM]{Fuzzy Cognitive Map}
 	\acro{sem}[SEM]{Structural Equation Model}
 	\acro{swig}[SWIG]{Single World Intervention Graph}
 	
 	\acro{pch}[PCH]{Pearl Causal Hierarchy}

 	\acro{sut}[SUT]{System Under Test}

    \acro{bn}[BN]{Bayesian Network}
    \acro{dag}[DAG]{Directed Acyclic Graph}
    \acro{scm}[SCM]{Structural Causal Model}

    \acro{adas}[ADAS]{Advanced Driver Assistance System}
    \acro{eba}[EBA]{Emergency Brake Assist}
 
\end{acronym}

\bibliography{bibliography.bib}


\end{document}